\documentclass[]{article} 
\usepackage{aaai21}  
\usepackage{times}  
\usepackage{helvet} 
\usepackage{courier}  
\usepackage[hyphens]{url}  
\usepackage{graphicx} 
\usepackage{natbib}  
\usepackage{caption} 
\usepackage{url}
\usepackage{makecell}
\usepackage{multirow}
\usepackage{subcaption}
\usepackage{amsmath}
\usepackage{bbm}
\usepackage{algorithmic}
\usepackage{amssymb}
\usepackage{bm}
\usepackage[ruled,vlined,linesnumbered]{algorithm2e}
\usepackage{enumitem}
\usepackage{pifont}
\usepackage{xcolor}
\usepackage{hyperref}
\usepackage{graphicx}

\title{Anticipating Driving Behavior through Deep Learning-Based Policy Prediction}
\author {
    Alexander Liu\\
}

\begin{document}

\maketitle
\begin{abstract}
In this project, we implemented an end-to-end system that takes in combined visual features of video frames from a normal camera and depth information from a cloud points scanner, and predicts driving policies (vehicle speed and steering angle). We verified the safety of our system by comparing the predicted results with standard behaviors by real-world experienced drivers. Our test results show that the predictions can be considered as accurate in at lease half of the testing cases (50\%~80\%, depending on the model), and using combined features improved the performance in most cases than using video frames only.
\end{abstract}

\section{Introduction}
Driving policy learning is one of the core topics in the research of autonomous driving. Researchers have successfully applied machine learning to the topic, and specifically, neural networks and computer vision techniques have been proved to be a promising solution to this task. There are mainly two categories of methods towards such this problem:
\begin{itemize}
    \item End-to-end systems, which learn and predict the driving behaviors based on current visual input and previous vehicle states, i.e., mapping pixels from a video steam directly to a series of driving policies, including speed, steering, throttle and brake actions, etc. Various previous work have demonstrated that neural networks can learn effectively in simple to moderate scenarios. 
    \item Non end-to-end systems, which usually involve further processing of the visual input, such as segmentation\cite{Babahajiani2017Urban3S}, object detection, or detecting a set of predefined affordance measurements, such as distance to surrounding cars \cite{chen2015deepdriving, li2023towards}, and such outputs are then feed into a rule-based system. By nature these intermediate outputs are much more human interpretable, but it can be hard to define a complete set of such affordance measurements, and we still need a lot of manual effort to associate these measurements with real-world scenarios. 
\end{itemize}
Safety is with no doubt the most important thing in autonomous driving technologies, and we need mechanisms to measure whether the self-driving system is safe to use or not. One solution is to verify the resemblance of driving behaviors between AI-prediction and human drivers (The assumption is experienced human drivers make reasonable decision and drive safely). So if there is a higher similarity between the predictions produced by self-driving systems and behaviors collected from human drivers, we consider the models intelligent and learning, and self-driving policies accurate.


\section{Related Work}
\begin{figure}[t]
    \centering
      \includegraphics[width=0.42\textwidth]{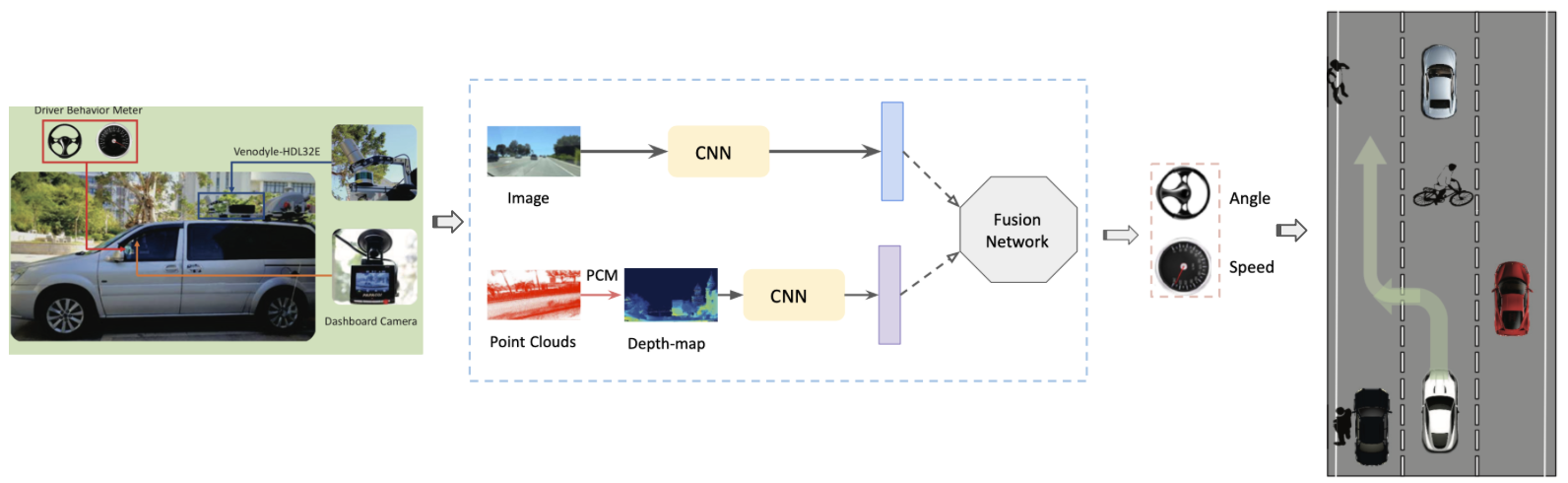}
    \caption{Workflow of our model. The first step is data collection, including images from video, point clouds LiDAR and driving behaviors. The driving behaviors are used as prediction values. The step second is the deep learning model proposed by us. The final step is the visualization of how the car runs.
    }
    \vspace{-0.2in}
    \label{fig:lead1}
\end{figure}
It has been understood that research and applications on autonomous driving behavior learning should leverage image and video cameras as well as laser scanners in order to comprehensively learn traffic and driving behavior. Deep learning techniques like neural networks are best positioned to address this problem because of the complexity of traffic.

Pomerleau et al. \cite{Pomerleau, liu2020visualnews, li2023towards} were the first researchers to use neural networks for lane following and obstacles avoiding. Since then, two primary methods for this task have been developed for solving this task \cite{xu2017end, liu2023documentclip}. These researchers also proposed a driving perplexity metric which is inspired by representative Markov model in linguistics.

In 2017, researchers at the Conference on Computer Vision and Pattern Recognition Workshops \cite{kim2017end, liu2020visual,liu2023aligning} showed that autonomous vehicles can learn driving policies in simple scenarios like highways. NVIDIA group used an end-to-end system, specifically a multi-layer convolution neural network, that maps directly from images for autonomous driving. Later, other researchers completed the same task for more complex driving scenarios.

However, these approaches only learn from large-scale videos because there were not sufficient precise laser-scanner data benchmarks. Chen et al. \cite{chendbnet, liu2023covid} were the first to propose a Driving Behavior Net (DBNet), which provides large-scale high-quality point clouds scanned by a Velodyne laser, videos recorded by a dashboard camera and standard drivers’ behaviors. These researchers confirmed that additional depth information improves networks’ abilities to determine driving policies. Nonetheless, it still has a large room to improve the usage of point cloud information.

\section{Method}
\begin{figure}[t]
    \centering
      \includegraphics[width=0.42\textwidth]{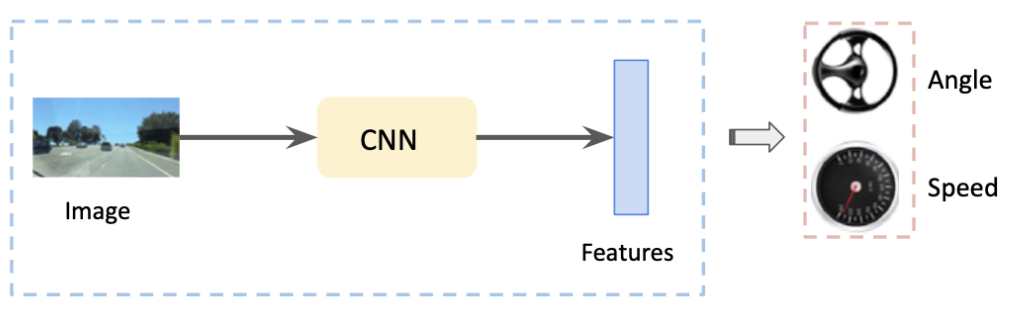}
    \caption{This is our basic model (IO), including one input: image. We use pretrained convolution neural network to extract the visual features from the images. After that a fully connected layer is leveraged to predict the speeds and angles.
    }
    \vspace{-0.2in}
    \label{fig:lead2}
\end{figure}
Generally speaking, we proposed an end-to-end deep learning network to predict the driving behaviors (Figure 2). In order to extract high-quality image features, we employ pre-trained models ResNet152 and Inception-v4. Different from \cite{chendbnet}, which simply used the concatenation operation to fuse the features before final prediction, we plan to leverage the more state-of-art techniques to thoroughly combine these features and discover the underlying connections in the future. What's more, as predicting the continuous values is much harder than the classification task, we will also try to find the extra guide mechanism to better predict the steering angles and vehicle speeds.
\begin{figure*}[t!]
    \centering
      \includegraphics[width=0.9\textwidth]{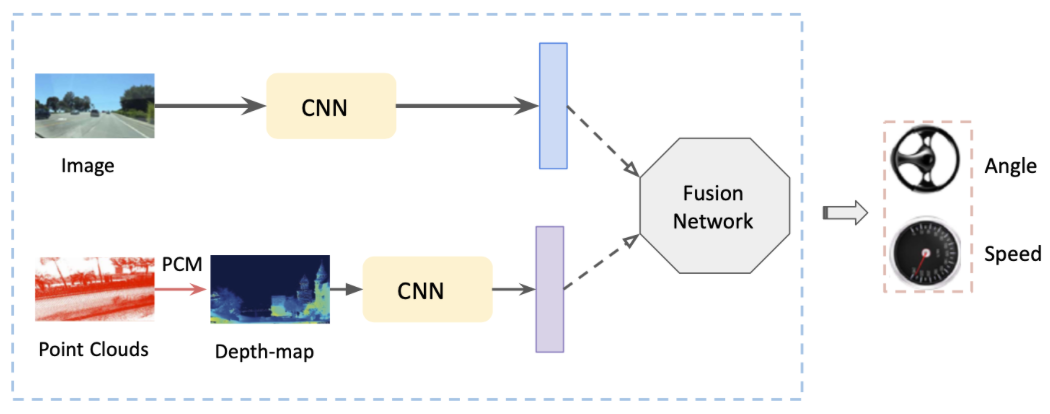}
    \caption{This is our advances model (PN), including two input: image and point clouds. As for the image input, we use pretrained convolution neural network to extract the visual features from the images. As for the point clouds, by using Point Clouds Mapping algorithm, we can get the depth-map and then another convolution neural network is utilized to extract to features. Finally, a fully connected layer is leveraged to fuse the two features and predict the speeds and angles.}
    \label{fig:model}
\end{figure*}

\begin{figure}[t]
    \centering
      \includegraphics[width=0.42\textwidth]{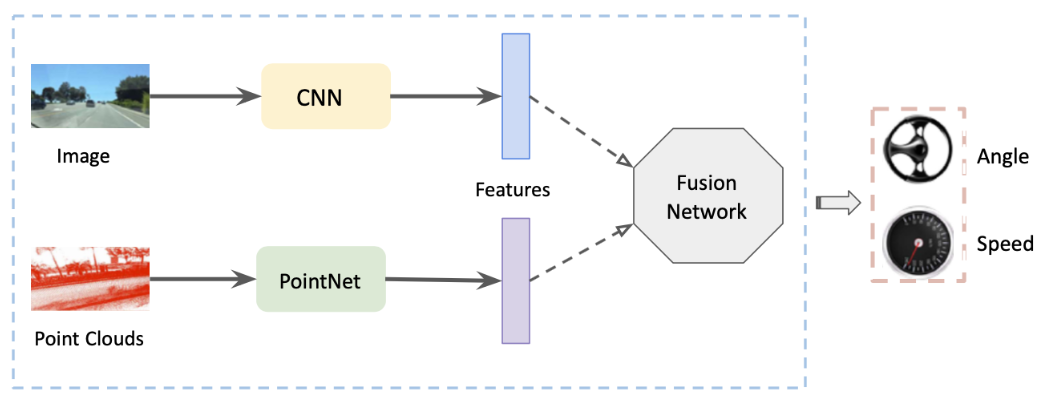}
    \caption{This is our advances model (PM), including two input: image and point clouds. As for the image input, we use pretrained convolution neural network to extract the visual features from the images. As for the point clouds, we utilze the PointNet to directly extract the geometry features from the disordered images . Finally, a fully connected layer is leveraged to fuse the two features and predict the speeds and angles.
    }
    \vspace{-0.2in}
    \label{fig:lead}
\end{figure}

\subsection{Targets}
As we known, the aim of our task is policy decision making in the field of automatic driving so as to improve the safety level just as a professional driver. In details, policy decision making process can be grouped into classification task with the decisions including turning right, turning left, going straight and stopping. It seems correct in the games but in the real-world driving, it's obviously not enough and too coarse to make sure the safe driving. In this case, continuous prediction is needed instead of discrete action prediction. As a result, vehicles are more likely to successfully drive by learning the regression task with angles and speed. That means the the inputs of the model are images from the digital video and cloud points from LiDAR. The outputs are according speeds and angles.

\begin{figure*}[t!]
    \centering
      \includegraphics[width=0.9\textwidth]{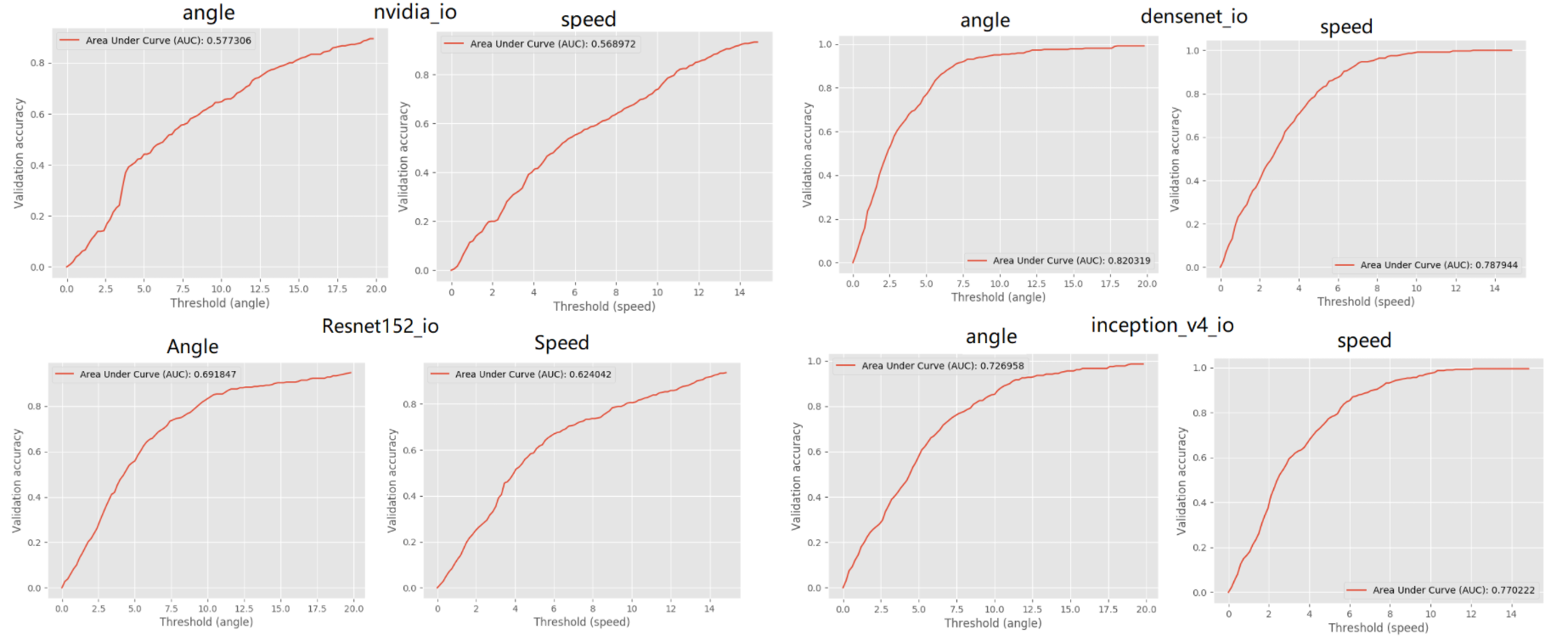}
    \caption{This is the experiment results of IO plot. IO means we only feeding images into the networks. We employed four different convolutional neural network: NVIDIA, ResNet-152, Incpetion-v4 and DenseNet. The experiment outputs are angles and speeds. The Details of of the model are in Figure 1.}
    \label{fig:io_plot}
\end{figure*}

\begin{figure*}[t!]
    \centering
      \includegraphics[width=0.9\textwidth]{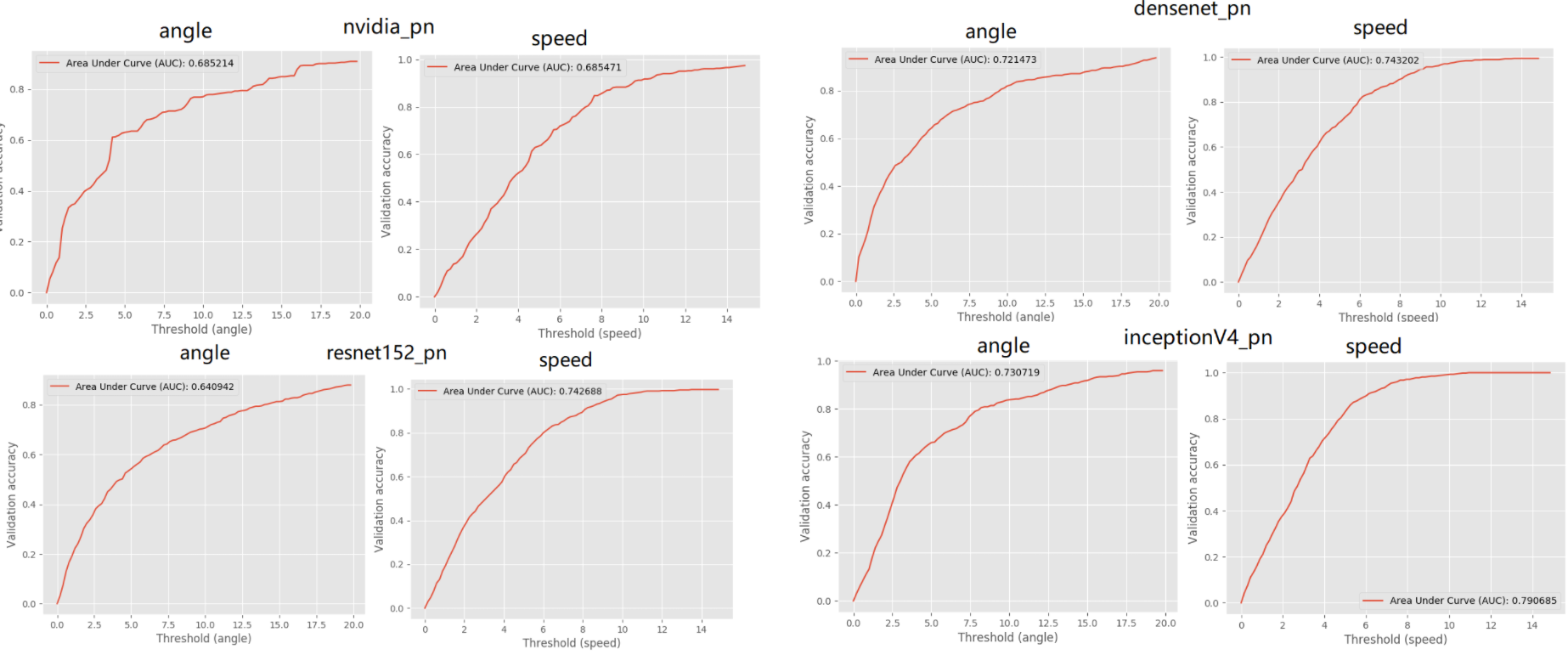}
    \caption{This is the experiment results of PN plot. PN denotes plain networks combined with PointNet architecture. We employed four different convolutional neural network: NVIDIA, ResNet-152, Incpetion-v4 and DenseNet. The experiment outputs are angles and speeds. The Details of of the model are in Figure 3.}
    \label{fig:pn_plot}
\end{figure*}

\subsection{Preliminary Components}
As illustrated in Figure 1, Figure 2 and Figure 3, inspired by many previous work~\cite{DBNet2018,agelastos2014lightweight}, we designed a series of models to predict the speed and angles. We have three versions of models in total, Image Only (IO), meaning that we only use image input, image plus Point Cloud Mapping (PCM) and image plus PointNet~\cite{qi2017pointnet}. The two latter models are both equipped with two inputs instead of one while the only difference is the ways they leveraged to process the cloud points.

In order to predict the speeds and angles, we are supposed to introduce the basic tools we use to extract the image features and depth features from the cloud components. As described in Figure 1, the first model we employ is CNN, where we adapt ResNet152~\cite{he2016deep}, Inception-v4~\cite{li2017image}. Both of the two models are state-of-art models in the field of computer vision, achieving perfect performance on the ImageNet~\cite{pennington2014glove} dataset. The model parameters here we used are pretrained ones from the ImageNet. Apart from this two, we also used NVIDIA architecture~\cite{kim2017end}, which is also the advanced model but much smaller than the previous two models. The CNN models are adapted in all the three versions of models. 

Another component is Points Clouds Mapping (PCM) targets at processing the
cloud points data. Specifically, it's able to nicely preserve the geometry information in the cloud points. By using PCM, we will get the Depth-map with colors containing the key features from the original data. After this, another convolution neural network is needed to extract the visual features. Instead of the two steps, we also tries another simple advanced network PointNet~\cite{qi2017pointnet}, which directly takes disorder points as the input of neural networks and finally output the representation features. The overall architecture of our model are shown in Figure 2 and Figure 3. What's more, the the comparison experiments will be shown in the experiment section.

\subsection{Models}
Generally speaking, we ran three models in total. The first one (Figure 1) is image only model, which follows the DNN end-to-end framework. This framework  receives one (or a mini-batch) frame input including images and predicts driving behavior. The RGB image feature extractor we used is Inception-v4, ResNet152 and NVIDIA. 

The Second model (Figure 2) absorbs images features from digital video and cloud points from LiDAR and then produce exact driving behaviors including speeds and angles. The RGB image feature extractor we used is Inception-v4, ResNet152 and NVIDIA. The Point Clouds feature extractor we used is Point Cloud Mapping(PCM). Then we will use another CNN to extract features. After getting the image features and cloud points features, a fusion network will be used to combine the two features. As for the second model in Figure 3, the only improvement is that we used a more efficient framework called PointNet to directly extract features from the cloud points.

We came up with several different ways to combine the features. The first one is the concentration operation, meaning we used a fully connected hidden layer to mix the the image and cloud points features and then predict the speeds and angles, which are trained with the loss functions together. This method is simple while the drawback is also obvious. So in order to make the model more complex, LSTM or attention mechanism could be employed. LSTM is able to leverage the time information to better predict the value and attention mechanism can learn the correlation between the two features and how to combine them more efficiently.

\section{Dataset}
\begin{figure}[h]
  \centering
  \includegraphics[width=\linewidth]{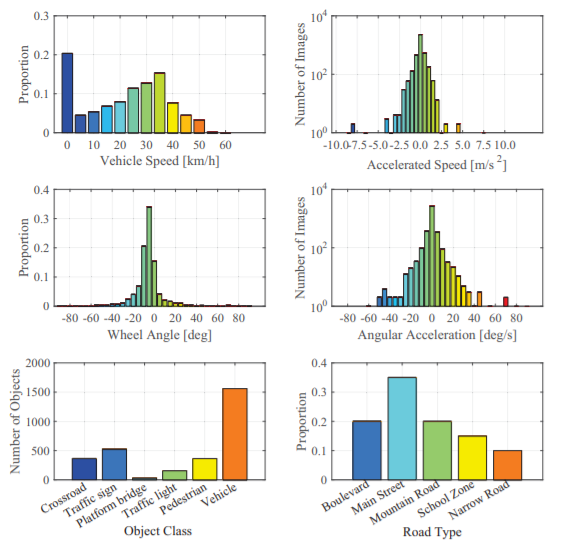}
  \caption{Driving Behavior Distribution and Object Statistics from DBNet dataset.}
  \label{fig:dataset_stats}
\end{figure}

We adapted the LiVi-Set from DBNet \cite{DBNet2018} as our training and testing dataset. It is a large-scale, high-quality dataset containing point clouds scanned by a Velodyne laser, driving videos recorded by a dashboard camera, and standard driving behaviors (vehicle speed and steering angle) collected from drivers with more than 10 years of driving experience using real-time sensors. All sensors are manually calibrated and data is synchronised. \newline \textbf{Features} LiVi-Set is the first dataset to combine 2D and 3D features and use depth information to make driving behavior predictions and is very different from other vision-based autonomous driving benchmarks. In addition, it covers a variety of driving scenarios such as traffic conditions (local route, boulevard, primary road, mountain road, school areas, etc), traffic volume (light, normal and heavy) and the number of pedestrians captured by the camera. Figure \ref{fig:dataset_stats} describes the diversity in details.
\newline \textbf{Preprocessing.} The driving videos are first down-sampled to 1 fps and then later reshaped to fit pre-trained neural networks (Nvidia takes 66$\times$200, ResNet and DenseNet takes 224$\times$224, and Inception takes 299$\times$299). The point clouds are also down-sampled to 16384x3 to fit in PointNet. However the feature maps are still generated from the original point clouds to keep as much features as possible. Furthermore, we only used ~10\% of the original dataset to speed up training and testing.  

\section{Experiments}
This section contains two parts: tasks of our research and evaluation methods we will leverage to analyse the performance our model.
\newline \newline
Generally speaking, our task can be divided into two sections, discrete prediction(classification task) and continuous prediction(regression task). As for the discrete prediction task, we aim to predict the current probability distribution over actions categories. To be more clear, the driving actions may include straight, stop, turn right, turn left and back. After this, we can decide whether it has safe issues. However, simply considering this policy making problem as a classification task is not suitable for the real driving since the actions are very limited, which will be dangerous for the drivers. We will design experiments to demonstrate its drawbacks. On the other hand, we will also employ multiple perception information including video frames and point clouds to predict steering angles and vehicle speeds.
\newline \newline
In order to evaluate the performance of driving behavior prediction by our model, we will use two metrics: perplexity metric and accuracy metric. Perplexity metric, proposed by \cite{xu2017end}, is the exponent of the sum of entropy in sequential prediction events. The prediction is more accurate if the perplexity metric is smaller than one meanwhile is larger than zero. Obviously, this metric is more suitable to represent the loss. As a result, we use accuracy metric, which uses a tolerance threshold to decide whether the case is correct. In detail, if the bias between the predictions and ground truth is smaller than the threshold, we will say the driving behavior is smooth and safe.


\section{Performance Evaluation}

\begin{figure}[h]
  \centering
  \includegraphics[width=\linewidth]{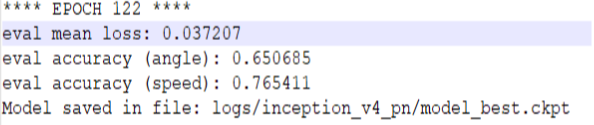}
  \caption{How a checkpoint is saved.}
  \label{fig:ckpt}
\end{figure}

We train models by using CentOS Linux 7 Server with one RTX 2080Ti Graphic Card. We use Python 2.7.1 and Tensorflow-gpu 1.11 as our software package tool. Each model was trained for 125 epochs and the we use the Root-Mean-Square Deviation (RMSD) = $\sqrt{MSE(\hat{\theta})} =\sqrt{E(\hat{\theta_1}- \hat{\theta_2} )^2} $ as our loss function.
\newline\newline
We select the best checkpoint if the difference between ground truth of angle value and predicted value of angles is less than $(5/180)*\pi=0.087$ ,and if the difference between ground truth of speed value and predicted value of speed is less than 0.25. The figure \ref{fig:ckpt}  shows the a checkpoint is saved as the best one.
\newline\newline
We set range of threshold to see the how accuracy of angle/speed grows with the increases og the threshold. Note that a higher threshold means that the difference between ground truth and predicted data is larger.  
Figure \ref{fig:io_plot} shows the performance or accuracy of our IO models for Nvidia, Densenet, Resnet and inceptionV4 network. It can be known from the plot, for the IO model, Densenet converges much faster than other networks. The angle accuracy and speed accuracy increase to around 90\% then threashold is set to 12.5 and 6 separately.

Figure \ref{fig:pn_plot} shows the performance or accuracy of our PN models for Nvidia, Densenet, Resnet and inceptionV4 network. It can be known from the plot, for the PN model, InceptionV4 converges much faster than other networks. Both angle accuracy and speed accuracy increase to around 90\% then threashold is set to around 7-8.

\begin{figure}[h]
  \centering
  \includegraphics[width=\linewidth]{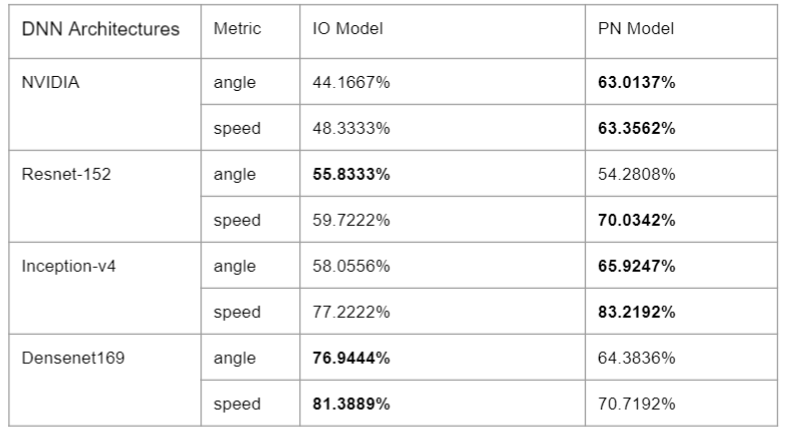}
  \caption{Experiment Performance. IO means we only feed image into our model. PN means we use both images and point clouds to predict the angles and speeds, meanwhile PointNet is leveraged to extract features from the points clouds.}
  \label{fig:res}
\end{figure}

Figure \ref{fig:res} shows the angle/speed accuracy of each IO Model and PN Model when threshold is set to 5. The accuracy of PN Models are better than IO Model in the most cases. The PN model with inception-V4 has the highest angle accuracy (65.9247\%) and highest speed accuracy(83.2192\%).
For densenet-169, the angle accuracy(76.9444\%) and speed accuracy (81.3889\%) of IO Model is much higher than PN model. 
\label{sec:Performance evaluation}

\section{Conclusion and Future Work}
In conclusion, our project implemented an end-to-end system and used the dataset that contains images and point cloud data to train IO and PM model. PM Model has higher accuracy than the IO Model. The angle/speed prediction is important in the real world as it can evaluating if a self-driving is safe or not. If the self-driving system provide the similar speed/angle result with the ground truth(i.e. data collected from the experienced drivers), we believe the system is safe to use. Due to the time limitation, we did not implement the PM model as well as LSTM. So in the future, we plan to finished them and automate the training and evaluating process.
\label{sec:conclusion}

\newpage
\bibliography{AAAI21}

\begin{thebibliography}{18}
\providecommand{\natexlab}[1]{#1}
\providecommand{\url}[1]{\texttt{#1}}
\providecommand{\urlprefix}{URL }
\expandafter\ifx\csname urlstyle\endcsname\relax
  \providecommand{\doi}[1]{doi:\discretionary{}{}{}#1}\else
  \providecommand{\doi}{doi:\discretionary{}{}{}\begingroup
  \urlstyle{rm}\Url}\fi

\bibitem[{Agelastos et~al.(2014)Agelastos, Allan, Brandt, Cassella, Enos,
  Fullop, Gentile, Monk, Naksinehaboon, Ogden
  et~al.}]{agelastos2014lightweight}
Agelastos, A.; Allan, B.; Brandt, J.; Cassella, P.; Enos, J.; Fullop, J.;
  Gentile, A.; Monk, S.; Naksinehaboon, N.; Ogden, J.; et~al. 2014.
\newblock The lightweight distributed metric service: a scalable infrastructure
  for continuous monitoring of large scale computing systems and applications.
\newblock In \emph{SC'14: Proceedings of the International Conference for High
  Performance Computing, Networking, Storage and Analysis}, 154--165. IEEE.

\bibitem[{Babahajiani et~al.(2017)Babahajiani, Fan, K{\"a}m{\"a}r{\"a}inen, and
  Gabbouj}]{Babahajiani2017Urban3S}
Babahajiani, P.; Fan, L.; K{\"a}m{\"a}r{\"a}inen, J.; and Gabbouj, M. 2017.
\newblock Urban 3D segmentation and modelling from street view images and LiDAR
  point clouds.
\newblock \emph{Machine Vision and Applications} 28: 679--694.

\bibitem[{Chen et~al.(2015)Chen, Seff, Kornhauser, and
  Xiao}]{chen2015deepdriving}
Chen, C.; Seff, A.; Kornhauser, A.; and Xiao, J. 2015.
\newblock Deepdriving: Learning affordance for direct perception in autonomous
  driving.
\newblock In \emph{Proceedings of the IEEE international conference on computer
  vision}, 2722--2730.

\bibitem[{Chen et~al.(2018)Chen, Wang, Li, Lu, Luo, HanXue, and
  Wang}]{DBNet2018}
Chen, Y.; Wang, J.; Li, J.; Lu, C.; Luo, Z.; HanXue; and Wang, C. 2018.
\newblock LiDAR-Video Driving Dataset: Learning Driving Policies Effectively.
\newblock In \emph{The IEEE Conference on Computer Vision and Pattern
  Recognition (CVPR)}.

\bibitem[{Chen et~al.()Chen, Wang, Li, Lu, Luo, Xue, and Wang}]{chendbnet}
Chen, Y.; Wang, J.; Li, J.; Lu, C.; Luo, Z.; Xue, H.; and Wang, C. ????
\newblock DBNet: A Large-Scale Dataset for Driving Behavior Learning .

\bibitem[{He et~al.(2016)He, Zhang, Ren, and Sun}]{he2016deep}
He, K.; Zhang, X.; Ren, S.; and Sun, J. 2016.
\newblock Deep residual learning for image recognition.
\newblock In \emph{Proceedings of the IEEE conference on computer vision and
  pattern recognition}, 770--778.

\bibitem[{Kim and Park(2017)}]{kim2017end}
Kim, J.; and Park, C. 2017.
\newblock End-to-end ego lane estimation based on sequential transfer learning
  for self-driving cars.
\newblock In \emph{Proceedings of the IEEE Conference on Computer Vision and
  Pattern Recognition Workshops}, 30--38.

\bibitem[{Li et~al.(2017)Li, Tang, Deng, Zhang, and Tian}]{li2017image}
Li, L.; Tang, S.; Deng, L.; Zhang, Y.; and Tian, Q. 2017.
\newblock Image caption with global-local attention.
\newblock In \emph{Thirty-First AAAI Conference on Artificial Intelligence}.

\bibitem[{Li et~al.(2023)Li, Xu, Liu, and Song}]{li2023towards}
Li, Z.; Xu, P.; Liu, F.; and Song, H. 2023.
\newblock Towards Understanding In-Context Learning with Contrastive
  Demonstrations and Saliency Maps.
\newblock \emph{arXiv preprint arXiv:2307.05052} .

\bibitem[{Liu et~al.(2023)Liu, Lin, Li, Wang, Yacoob, and
  Wang}]{liu2023aligning}
Liu, F.; Lin, K.; Li, L.; Wang, J.; Yacoob, Y.; and Wang, L. 2023.
\newblock Aligning Large Multi-Modal Model with Robust Instruction Tuning.
\newblock \emph{arXiv preprint arXiv:2306.14565} .

\bibitem[{Liu, Tan, and Tensmeyer(2023)}]{liu2023documentclip}
Liu, F.; Tan, H.; and Tensmeyer, C. 2023.
\newblock DocumentCLIP: Linking Figures and Main Body Text in Reflowed
  Documents.
\newblock \emph{arXiv preprint arXiv:2306.06306} .

\bibitem[{Liu et~al.(2020{\natexlab{a}})Liu, Wang, Wang, and
  Ordonez}]{liu2020visual}
Liu, F.; Wang, Y.; Wang, T.; and Ordonez, V. 2020{\natexlab{a}}.
\newblock Visual news: Benchmark and challenges in news image captioning.
\newblock \emph{arXiv preprint arXiv:2010.03743} .

\bibitem[{Liu et~al.(2020{\natexlab{b}})Liu, Wang, Wang, and
  Ordonez}]{liu2020visualnews}
Liu, F.; Wang, Y.; Wang, T.; and Ordonez, V. 2020{\natexlab{b}}.
\newblock Visualnews: Benchmark and challenges in entity-aware image
  captioning.
\newblock \emph{arXiv preprint arXiv:2010.03743} .

\bibitem[{Liu, Yacoob, and Shrivastava(2023)}]{liu2023covid}
Liu, F.; Yacoob, Y.; and Shrivastava, A. 2023.
\newblock COVID-VTS: Fact Extraction and Verification on Short Video Platforms.
\newblock \emph{arXiv preprint arXiv:2302.07919} .

\bibitem[{Pennington, Socher, and Manning(2014)}]{pennington2014glove}
Pennington, J.; Socher, R.; and Manning, C.~D. 2014.
\newblock Glove: Global vectors for word representation.
\newblock In \emph{Proceedings of the 2014 conference on empirical methods in
  natural language processing (EMNLP)}, 1532--1543.

\bibitem[{Pomerleau(1989)}]{Pomerleau}
Pomerleau, D.~A. 1989.
\newblock ALVINN: An Autonomous Land Vehicle in a Neural Network.
\newblock In \emph{Advances in neural information processing systems},
  305--313.

\bibitem[{Qi et~al.(2017)Qi, Su, Mo, and Guibas}]{qi2017pointnet}
Qi, C.~R.; Su, H.; Mo, K.; and Guibas, L.~J. 2017.
\newblock Pointnet: Deep learning on point sets for 3d classification and
  segmentation.
\newblock In \emph{Proceedings of the IEEE conference on computer vision and
  pattern recognition}, 652--660.

\bibitem[{Xu et~al.(2017)Xu, Gao, Yu, and Darrell}]{xu2017end}
Xu, H.; Gao, Y.; Yu, F.; and Darrell, T. 2017.
\newblock End-to-end learning of driving models from large-scale video
  datasets.
\newblock In \emph{Proceedings of the IEEE conference on computer vision and
  pattern recognition}, 2174--2182.

\end{thebibliography}

\clearpage

\end{document}